\documentclass[12pt, titlepage, reqno]{article}

 \usepackage[cmex10]{amsmath}

 \usepackage{amssymb, latexsym}
 \usepackage{amsmath} 
 \usepackage{amsthm}
 \usepackage{bm}
 \usepackage[mathscr]{eucal}
 \usepackage{enumerate}

 \usepackage{natbib}
 \usepackage[french,english]{babel}

\usepackage{algorithm,algpseudocode}
\usepackage{caption}
\usepackage{color}

\usepackage{gensymb}

\usepackage[caption = false]{subfig}

\usepackage{float}

\usepackage{algpseudocode}

\usepackage{graphicx}

\usepackage{tabularx}
\usepackage{slashbox}

\usepackage{hyperref}
\hypersetup{
    colorlinks,%
    citecolor=blue,%
    filecolor=blue,%
    linkcolor=blue,%
    urlcolor=blue,
}  
  
\usepackage{multirow}

\graphicspath{{./Figures/}}

\begin{document}

 \begin{titlepage}

 \begin{center}

\textbf{Bi-class classification of humpback whale sound units against complex background noise with Deep Convolution Neural Network}

\vspace{10ex}

Cazau Dorian$^{a}$ \footnote{Corresponding author e-mail: cazaudorian@outlook.fr}, Riwal Lefort$^{a}$, Julien Bonnel$^{a}$, Jean-Luc Zarader$^{b}$ and Olivier Adam$^{c}$\\ 

\begin{flushleft}

\small{$^a$ ENSTA Bretagne, Lab-STICC (UMR CNRS 6285), 2 rue Fran\c{c}ois Verny, 29806 Brest Cedex 09, France} \\
\small{$^b$ Inst. Syst. Intelligents et de Robotique, UPMC, Paris, France} \\
\small{$^c$ Sorbonne Universit\'es, UPMC, UMR 7190, Institut Jean Le Rond d'Alembert, Paris, France} 

\end{flushleft}

 \end{center}

 \end{titlepage}

\begin{abstract}
Automatically detecting sound units of humpback whales in complex time-varying background noises is a current challenge for scientists. In this paper, we explore the applicability of Convolution Neural Network (CNN) method for this task. In the evaluation stage, we present 6 bi-class classification experimentations of whale sound detection against different background noise types (e.g., rain, wind). In comparison to classical FFT-based representation like spectrograms, we showed that the use of image-based pretrained CNN features brought higher performance to classify whale sounds and background noise. 
\end{abstract}

\addtocounter{page}{2}

\section{Introduction}

In the frequency band from few tens of Hz up to 50 kHz, dominant sources of ambient noise in the ocean can be broadly divided into sounds resulting from geophony (i.e., sounds from natural physical processes, e.g. wind-driven waves), biophony (i.e. sounds from biological activities, e.g., whale vocalizations), and anthropophony (i.e. man-made sounds, e.g. commercial shipping) \citep{Wenz1962}. Among them, humpback whales produce complex stereotyped songs during the winter-spring breeding season. In the bioacoustic community, there is an important need to objectively and systematically detect, and further classify, the unit constituents of a song. However, some major difficulties exist in this task. Humpback whale songs are most often recorded in ``hot spot" gathering many whale individuals that sing altogether, making the identification of constitutive sounds of each song very difficult. In addition to these overlapped sounds, environmental and anthropogenic sources also corrupt song sound units. Also, the acoustic richness and variability of humpback whale sound units make them difficult to classify into robust discrete categories.

In this paper, we are interested in extracting vocalizations of singing whales from a complex time-varying marine environment. Most computational tools (e.g. PAMguard \cite{Gillespie2008}, Ishmael \cite{Mellinger2001}) for this task are based on hand-engineered features optimized for one specific source (e.g. a single whale in a given geographical location). However, in highly time-variable marine environments, the extraneous complexities of the ocean ambience prevent the feature engineering methods to capture the invariant features. In this paper, we explore the applicability of the deep learning method \citep{Hinton2006} to the task of identifying properly the sound units of a singing humpback whale within a complex multi-source underwater soundscape (composed of real acoustic sources, e.g. rain, wind, marine traffic and song chorus). The deep learning approach, with architectural resemblance to the mammalian neocortex, is deemed to capture the useful information hidden deeply in the actual observation by automatically learning features in a hierarchical manner. We are interested in quantifying the performance of image-based pretrained Convolution Neural Network (CNN) features to classify spectrograms (FFT-based features) of whale sounds and background noise (equivalent to a detection task), in comparison to classical FFT-based representation.

\section{Methods}\label{ResulAndDi}

\paragraph{Processing chain}

Figure \ref{SchemeBlock} shows the block diagram of our classification task, with details provided in the appendix. The input of this processing chain is a 2-s long temporal slice $m$ of a time series $x^{(m)}$ ($T$=2 s / 171 samples at $f_s$=44.1 kHz). This time series window is Fast-Fourier transformed into a spectrogram ($F$ = 2048 bins). These spectrograms $S(f,t)$ are then converted into 256 $\times$ 256 pixels .jpg images in uint-8 format. Each image is used as input of a pre-trained CNN. CNN models such as AlexNet, GoogLeNet that are pre-trained on ImageNet can indeed been used as generic feature extractors for various tasks and data types. This is done by removing the top output layer and using the activations from the last fully connected layer (CNN codes) as features. Pre-trained CNNs from the Vlfeat project\footnote{Website link: \url{http://www.vlfeat.org/matconvnet/pretrained}} were used. We tried the following networks from the imagenet framework \citep{Chatfield2014}: imagenet-vgg-f, imagenet-vgg-m, imagenet-vgg-s. After several experimentations, imagenet-vgg-f showed the best performance and was consequently used in our study. The output dimension $O_w$ of the CNN features $W^{(m)}$ is set to 4096. Eventually, the output result of our processing chain is a binary decision $D^{(m)}$ classifying this time series slice into a whale sound (i.e. 1) or background noise (i.e. 0). The two-class linear SVM used in this study is from LibLinear\footnote{Website link: \url{https://www.csie.ntu.edu.tw/~cjlin/liblinear/}} \citep{Fan2008}. Default parameters of the toolbox have been used, with in particular a radial basis function $e^{(-\gamma*|u-v|^2)}$ for the kernel of degree 3. We did not try to optimize such parameters for better classification performance.

\begin{figure}[htbp]
\centering
\includegraphics[width=\columnwidth]{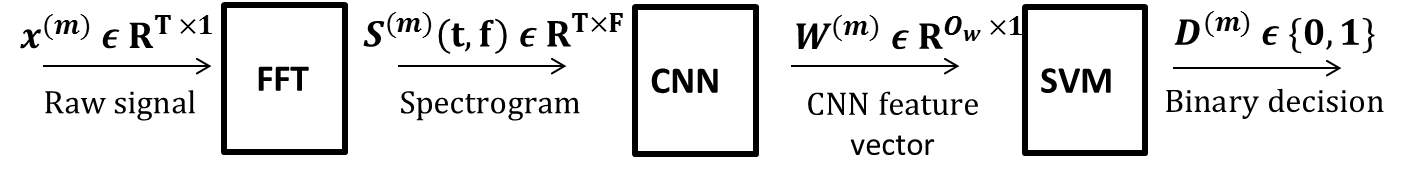}
\caption{Block diagram of our classification task.}\label{SchemeBlock}
\end{figure}

\paragraph{Evaluation dataset and procedure}

We first selected three ``clean" songs from different humpback whale singers in the Sainte Marie Island Channel, North East Madagascar (16 \degree S , 49 \degree E), during the month of August. These song recordings were done close to the singers (closer than 200 m) to improve signal-to-noise ratio. Great care was taken to record only the singing focal animal, in order to prevent any confounding effect of background noise and overlapping vocalizations. In each of the three songs, 500 sound units were manually selected through a visual inspection of the spectrogram outcomes under Adobe Audition software, for a total of 1500 sound units. We also collected data from four different background noise types: wind (geophony), rain (geophony), marine boat traffic (anthropogeny) and humpback whale song chorus (biophony). Full details are provided in the appendix. Evaluation datasets were built automatically by combining sound units of the focal whales with a noise sample taken successively from the different noise background types. Each sound unit file was first unitary normalized, and buffered into non-overlapped 2-s time windows. Each time window was then added to some background noise, randomly selected from the 5 background types, respecting a given Signal-to-Noise Ratio (SNR). This SNR was defined as a sensitivity parameter for our numerical experimentations, ranging from -10 dB to 10 dB. SNR is defined as $SNR = 10 log_{10} \frac{<x_s^2>}{<x_n^2>}$, where $<x_s^2> = \frac{1}{T} \int_0^T <x_s^2>^2(t) dt$ and where $x_s$ represents the recorded pressure of the sound unit time series, $x_n$ the background noise, and T= 2s is the duration of the signal. Note that negative SNR in the time domain does not imply negative SNR for individual frequencies following a transformation into the Fourier domain. As detailed in table \ref{Tab_backgroundTypes}, we designed 6 different numerical experimentations based on the dataset of sound units from the focal whale, combined with the previous evaluation background types. The sixth evaluation dataset was built by randomly selecting different noise samples from all possible background types, similarly to a one-against-all classification scenario.

\begin{table}[h]
\centering
\begin{tabular}[width=0.75\columnwidth]{|c|c|}
\hline

Experimentation names & Evaluation background types \\\hline

$E_1$ & Clean background \\\hline
$E_2$ & Wind \\\hline
$E_3$ & Rain \\\hline
$E_4$ & Marine traffic  \\\hline
$E_5$ & Song chorus  \\\hline
$E_6$ & Wind + Rain + Marine traffic + Song chorus\\\hline

\end{tabular}
\vspace{0.2cm}
\caption{Details of numerical experiments.}\label{Tab_backgroundTypes}
\end{table}

Eventually, our numerical experimentations consist of a two-class classification task, that aims to detect focal whale sound units against noise background. Training and test datasets were defined with sound samples from different years, and were set to 300 and 200 samples, respectively. For each year couple, final classification accuracy resulted from the average scores provided by a Monte Carlo simulation with 100 iterations. We then also averaged over years. Results are then displayed as 2D confusion matrices of Whale sound vs Noise background.

\section{Results and Discussions}

\begin{figure}[htbp]
\centering
\includegraphics[width=0.7\columnwidth]{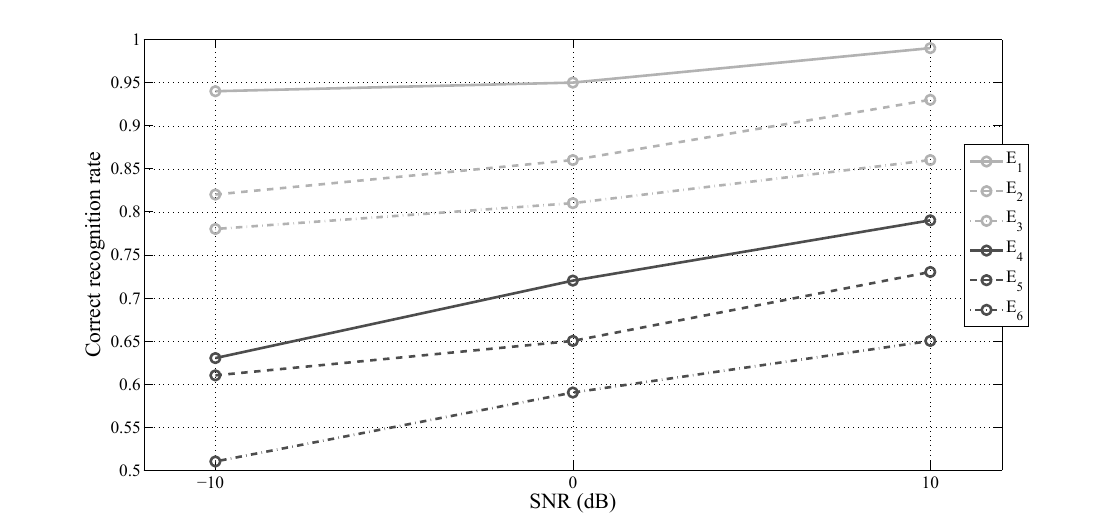}
\caption{Correct recognition rates against SNR for each experiment, i.e. corresponding to the different background noise types.}\label{CorrectRecSNR}
\end{figure}

Figure \ref{CorrectRecSNR} shows the dependency of correct recognition rates of whale sound units on SNR for the different background noise types. Supplementary results on the bi-class classification (with false negative rates) and on a comparative study between different detection systems are also provided in appendix. Experiment $E_1$ is the baseline performance, with a correct recognition rate of 95 \%, and a false alarm rate of 1\%. All other background noise types that have been used to corrupt whale sounds decreased baseline performance, from - 9 \% to -36 \% in the correct recognition rate, and from + 3 \% to + 36 \% in the false alarm rate.

Environmental noises like wind and rain had the smallest impact on classification performance. These acoustic sources have broadband frequency spectra, with a rather uniform distribution of the energy in the spectrum \citep{Nystuen2015}, which does not distort that much whale sound spectra. On the contrary, marine traffic (especially at short distance from the hydrophone) generates a great variety of sound spectra that are more similar to whale sounds, with most often a time-varying harmonic content. Naturally, song chorus is the background noise type that decreases the more classification performance, as it is composed of a multitude of whale sounds from different singers. Globally, decrease in SNR between whale sounds and background noise degrades classification performance. Variations in SNR appear to affect to a higher extent performance of experiments $E_4$ and $E_5$. As already commented, the acoustic content of these noise types are more similar to whale sounds, and consequently lower SNR values reinforce ambiguity between them.

\section*{Appendix}

\subsubsection*{Acknowledgments}

The authors also would like to thank the Direction G\'en\'erale de l'Armement (DGA, France) for supporting this work. It was also financially supported by the Dirac association (Paris, France).

\subsubsection*{Theoretical background and implementation details for each processing block}

\paragraph{FFT}

The power spectrum is used to measure the amplitude of discrete frequency components. The power spectra for each of the data segments $t$ are then combined to form an array in frequency and time $S^{(m)}(t,f)$ that we call spectrogram, 

\begin{equation}
S^{(m)}(t,f) = 10 log_{10} \frac{| X_t^{(m)}(f) |}{p_{ref}^2}
\end{equation}

where $X_t^{(m)}(f) = \sum_{n=0}^{N-1} x_t^{(m)}[n] e^{\frac{-i 2\pi f n}{N}}$ is the FFT transformed of the time series signal $x$, and $p_{ref}$ is a reference pressure of 1 $\mu$Pa. The spectrogram is then used to measure the amplitude of discrete frequency components against time. Spectrogram parameters: segment size: 22 ms (1024 samples with a sample frequency $f_s$=44.1 kHz) ; 50\% overlap (temporal resolution: 11 ms) ; F = 2048 bins (i.e. FFT size, providing a spectral resolution of 21 Hz) ; Hamming window.

\paragraph{CNN}

CNN is a type of artificial neural networks inspired by visual information processing in the brain. To recognize complex features from the visual information, CNN consists of several layers which extract and repeatedly combine low-level features for the composition of high-level features. The composed high-level features are used for CNN to classify an input image. In our processing chain, FFT-based spectrograms $S(f,t)$ were converted into 256 $\times$ 256 pixels .jpg images in uint-8 format. Then, each image is used as input of the pre-trained CNN, giving a feature vector called the CNN code. For the process and the simplicity of computation, many CNN structures are described by combinations of convolutional, pooling, and fully-connected layers. Convolutional layer (C-layer) extracts higher-level features by convolving received feature maps from the previous layer and activating the convolved features. By supposing that j-th layer has $N_j$ nodes (neurons) whose output feature map is $h_i^j$, then :

\begin{equation}
h_i^j = g ( \sum_{n=1}^{N_{j-1}} \bm{\phi_{in}} * h_n^{j-1} + b_i^j)
\end{equation}

for $j=1,...,N_j$, where $\bm{\phi_{in}}$ is the convolutional filter connecting i-th node (neuron) on the $j^{th}$ layer and n-th node (nueron) on the j − 1-th layer, $b_i^j$ is the bias for the i-th output on the $j^{th}$ layer, and f is the activation function. Given its efficiently of computation and common use, ReLU $g(x) = max(0, x)$ is the used activation function. A C-layer usually is followed by a pooling layer (P-layer) which reduces the dimension of feature maps by ``max pooling". The max pooling downsamples the input feature maps by striding a rectangular receptive field and taking the maximum in the field. For example, we use 2 $\times$ 2 receptive field with stride 2 in our research, which reduces the dimension of feature map by 1/4. After couples of pairs of C-layer and P-layer, a fully-connected layer (F-layer) integrates high-level features and produces compact feature vectors:

\begin{equation}
h_i^j = g ( \sum_{n=1}^{N_{j-1}} < \bm{\phi_{in}} , h_n^{j-1} + b_i^j)
\end{equation}

for the $j^{th}$ F-layer. Note that we use inner-product rather than convolution between the filter Win and the input feature map $h_n^{j-1}$. Like the C-layers, ReLU is used as the activation function of the F-layers. On the final layer, say J-th layer, the output layer produces the posterior probability for each class by the softmax function. The idea of exploring CNN features is motivated by their usefulness on a wide variety of tasks. The activations which are the output $W^{(m)}$ of CNN layers can be interpreted as visual features. 

\paragraph{SVM}

Given a set of training examples of CNN features, each marked as belonging to whale sounds or background noise, an SVM training algorithm builds a model that assigns new examples from a test dataset to one category or the other, making it a non-probabilistic binary linear classifier  \citep{Fan2008}.

\subsubsection*{Details on background noise types}

\paragraph{Clean background}
Clean background data were extracted from the three songs used for sound unit extraction, and correspond to the time intervals between each sound unit manually labelled. As already mentioned, these song recordings were performed in clean meteorological conditions ans without observed external sources.

\paragraph{Wind}
Wind-driven background data samples were extracted from two different underwater sound datasets recorded in shallow water environments. We only collected wind events belonging to the see state of Beaufort scale 6 or higher (identified to high and gale winds with a speed higher than 7 m.s$^{-1}$) \citep{Wenz1962}. These wind speed values were extracted from a moored meteorological buoy in Saint-Pierre-et-Miquelon (from the National Climatic Data Center\footnote{Website link: \url{https://www.ncdc.noaa.gov/}}) and the European Center for Medium range Weather Forecasting (ECMWF) analysis, that provide gridded daily-averaged wind and wind stress fields over global oceans through satellite imaging and assimilation models \citep{Bentamy2011}.

\paragraph{Rain}
As for wind, rain-driven background data samples were extracted from the Saint-Pierre-et-Miquelon and the Sainte-Marie channel. We also only collected rain events superior to 10 $mm/h$, that have already been proved to have a significant effect on ambient noise \citep{Pensieri2015}. These precipitation rates were extracted from a moored meteorological buoy in Saint-Pierre-et-Miquelon (from the National Climatic Data Center\footnote{Website link: \url{https://www.ncdc.noaa.gov/}}) and the European Center for Medium range Weather Forecasting (ECMWF) analysis. 

\paragraph{Marine boat traffic}
Our data samples for background noise from marine boat traffic were extracted from the Saint-Pierre-et-Miquelon dataset in recordings from 19/08/2010 to 02/11/2010 and in the Sainte Marie channel. Most encountered ships are motor boats (from 15m to 25m long) used for material shipping or passenger transport. Ship tracklines around the recording hydrophone ($\approx$ 40 km$^2$ searching area) were taken from the World Meteorological Organization Voluntary Observing Ship Scheme (VOS) climate project \footnote{Download link: \url{http://www1.ncdc.noaa.gov/pub/data/vosclim/}}. This project is based on ships of many countries that voluntarily transmit their location updates multiple times a day under this program. 

\paragraph{Song chorus}
Eventually, song chorus were extracted from the Sainte Marie channel (Madagascar), from other recordings and time periods of the songs than the ones used previously.

\subsubsection*{Supplementary results}

\begin{figure}[htbp]
\centering
\includegraphics[width=0.98\columnwidth]{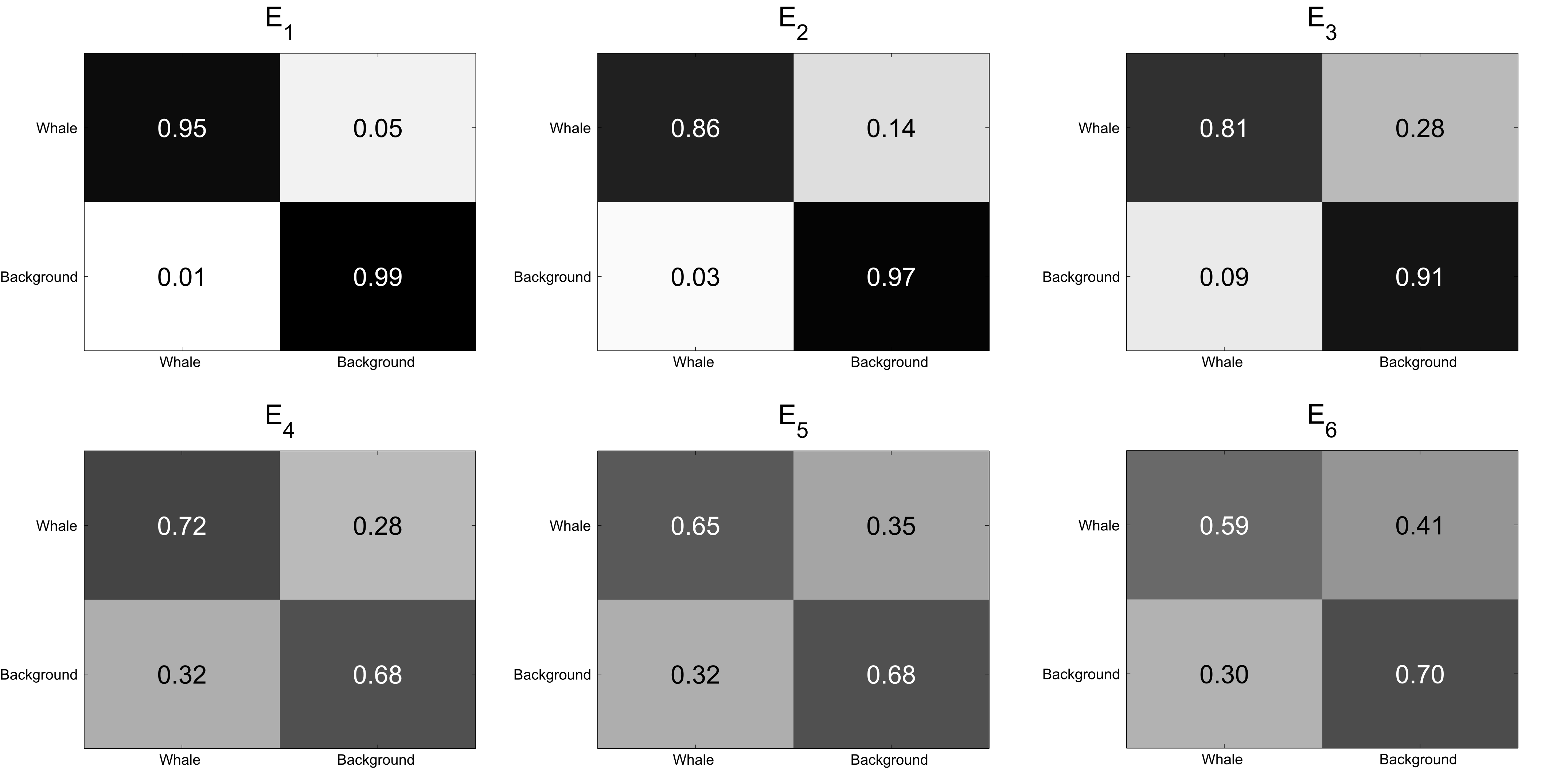}
\caption{Confusion matrices of the experiments $E_1$ to $E_6$.}\label{confusion_matrix}
\end{figure}

\begin{figure}[htbp]
\centering
\includegraphics[width=0.98\columnwidth]{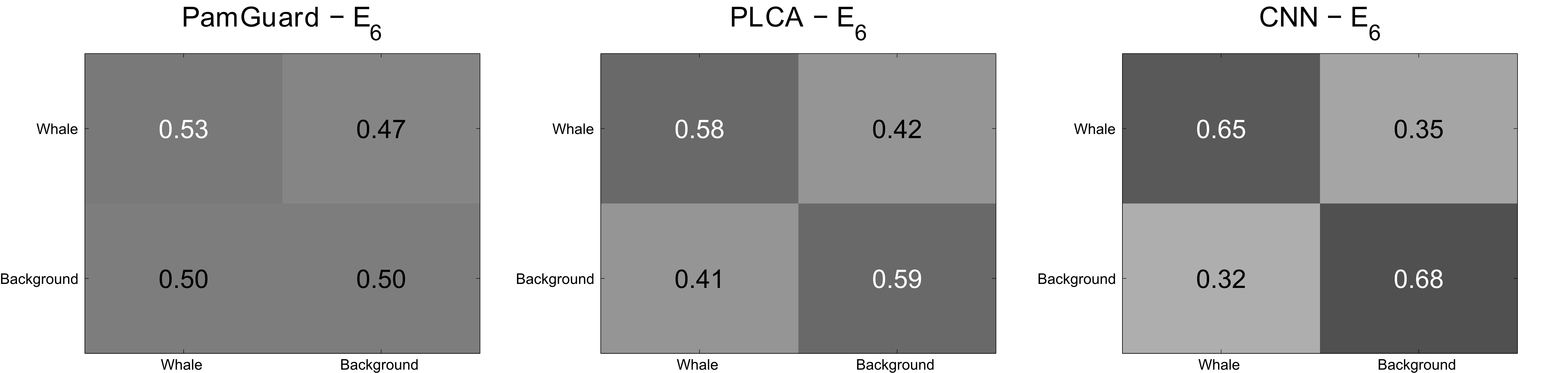}
\caption{Confusion matrices of the experiments $E_6$ with different classification systems: CNN, PLCA and PamGuard.}\label{confusion_matrix2}
\end{figure}

Figure \ref{confusion_matrix} represents confusion matrices of the experiments $E_1$ to $E_6$, where a SNR of 0 dB was set. We also compared detection performance of our method with two other systems: PamGuard\footnote{Publically available sofwtare that has been widely used in the marine mammal community for the task of vocal sound detection in particular. Downloading links of this software is \url{http://www.pamguard.org/}.} \cite{Gillespie2008} and a PLCA-based system that shows promising performance for humpback whale sound detection \citep{Cazau2016}. Figure \ref{confusion_matrix2} represents confusion matrices of the experiments $E_6$, where a SNR of 10 dB was set. In comparison to classical FFT-based representation, we observe that the use of image-based pretrained CNN features brought higher performance to classify spectrograms (FFT-based features) of whale sounds and background noise, with gains in correct classification rates of + 12 \% and + 7 \%, in reference to PamGuard and PLCA systems, respectively.

\end{document}